\documentclass{sigchi-ext}
\usepackage[T1]{fontenc}
\usepackage{textcomp}
\usepackage[scaled=.92]{helvet} 
\usepackage{graphicx} 
\usepackage{balance}  
\usepackage{booktabs} 
\usepackage{ccicons}  
\usepackage{ragged2e} 

\usepackage{algorithmic}
\usepackage{algorithm}
\usepackage{tabularx}
\usepackage{comment}
\usepackage{booktabs}
\usepackage{enumitem}

\def\plaintitle{Exploring Apprenticeship Learning for Player Modelling in Interactive Narratives} \def\plainauthor{Jessica Rivera-Villicana, Fabio Zambetta, James Harland, Marsha Berry}

\def\plainkeywords{Player Modelling; Apprenticeship Learning; Interactive Narratives;
  Inverse Reinforcement Learning; Anchorhead.}

\title{Exploring Apprenticeship Learning for Player Modelling in Interactive Narratives}

\numberofauthors{4}
\author{%
  \alignauthor{%
    \textbf{Jessica Rivera-Villicana\footnote{During the development of this work, the author was a PhD candidate in the School of Science, RMIT University.}}\\
    \affaddr{Applied Artificial Intelligence Institute (A\textsuperscript{2}I\textsuperscript{2})} \\
    \affaddr{Deakin University} \\
    \affaddr{Burwood, VIC, Australia}\\
    \email{riverav@deakin.edu.au} }\alignauthor{%
    \textbf{Fabio Zambetta}\\
    \affaddr{School of Science}\\
    \affaddr{RMIT University}\\
    \affaddr{Melbourne, VIC, Australia}\\
    \email{fabio.zambetta@rmit.edu.au} } \vfil \alignauthor{%
    \textbf{James Harland}\\
    \affaddr{School of Science}\\
    \affaddr{RMIT University}\\
    \affaddr{Melbourne, VIC, Australia}\\
    \email{james.harland@rmit.edu.au} }\alignauthor{%
    \textbf{Marsha Berry}\\
    \affaddr{School of Media and Communication}\\
    \affaddr{RMIT University}\\
    \affaddr{Melbourne, VIC, Australia}\\
    \email{marsha.berry@rmit.edu.au}}}

\definecolor{linkColor}{RGB}{6,125,233}
\hypersetup{%
  pdftitle={\plaintitle},
  pdfauthor={\plainauthor},
  pdfkeywords={\plainkeywords},
  bookmarksnumbered,
  pdfstartview={FitH},
  colorlinks,
  citecolor=black,
  filecolor=black,
  linkcolor=black,
  urlcolor=linkColor,
  breaklinks=true,
}

\setcopyright{rightsretained}
\begin{document}

\conferenceinfo{Extended Abstracts of the 2019 Annual Symposium on Computer-Human Interaction in Play}{October 22--25, 2019, Barcelona, Spain}
\doi{https://doi.org/10.1145/3341215.3356314}
\isbn{ISBN 978-1-4503-6871-1/19/10}
\copyrightinfo{\acmcopyright}

\maketitle

\RaggedRight{} 

\begin{abstract}
    In this paper we present an early Apprenticeship Learning approach to mimic the behaviour of different players in a short adaption of the interactive fiction Anchorhead. Our motivation is the need to understand and simulate player behaviour to create systems to aid the design and personalisation of Interactive Narratives (INs). INs are partially observable for the players and their goals are dynamic as a result. We used Receding Horizon IRL (RHIRL) to learn players' goals in the form of reward functions, and derive policies to imitate their behaviour. Our preliminary results suggest that RHIRL is able to learn action sequences to complete a game, and provided insights towards generating behaviour more similar to specific players. 
	
\end{abstract}
\graphicspath{{Imgs/}}



\section{Introduction}
We present an Apprenticeship Learning (AL) approach to Player Modelling (PM) in Interactive Narratives (INs), with the specific focus of replicating players behaviour within the game.

\marginpar{%
  \fbox{%
    \begin{minipage}{\marginparwidth}
\textbf{Research aims: }
\begin{enumerate}
\item{To capture human decision making, which can result in sub-optimal behaviour based on the information available to players at a given game state.}
\item{To mimic specific behaviour for players with different preferences and gaming styles.}
\end{enumerate}
 \end{minipage}}\label{sec:researchaims} }

An Interactive Narrative (IN) is a form of storytelling where the player can choose from different options at a given point and progress throughout the story depending on their choice. As INs gain popularity in commercial computer games (e.g., The Witcher~\cite{CDProjectRED}, The Elder Scrolls~\cite{Studios} or Heavy Rain~\cite{Dream2010}), the interest in knowledge about the player increases, with the aim to deliver experiences that meet their preferences.



This work consists on formulating the problem of emulating the behaviour of human players within an IN as an AL via Inverse Reinforcement Learning (IRL) problem. By taking a set of game traces generated by different players to learn a reward function that captures their goals without manually defining them, we aim to be able to reproduce their behaviour with an artificial agent.


\marginpar{%
  \vspace{-5pc} \fbox{%
    \begin{minipage}{\marginparwidth}
     \textbf{Anchorhead} is an interactive fiction written by Michael S. Gentry in 1998~\cite{anchorhead}. The story takes place in the fictional town of Anchorhead, where the protagonist and her husband inherited a mansion from distant family. The previous owner of the house, Edward Verlac, killed his family and committed suicide. To reach one of the two endings of the extract used in this work, the player needs to explore the world by visiting places, examining objects, talking to characters, etc. to find out the truth behind the Verlac family.
    \end{minipage}}\label{sec:anchorhead}}

We consider AL an appropriate approach to simulate players' behaviour in INs for the following reasons:\\
1) An IN environment is only partially observable for the player, and the game's goals differ depending on the story and the elements discovered so far. This makes the player's behaviour sub-optimal and their goals dynamic; a player working on a quest does not always mean they will finish it before working on a new one because they constantly set, re-define and re-prioritise their goals as they learn new information~\cite{Baikadi2013}.\\
2) Besides the sub-optimality of player behaviour due to their partial knowledge of the IN's world, the behaviour of each player is affected by their individual preferences and gaming style. For instance, a player who likes collecting items is more likely to take objects as they find them, regardless of their potential use in the game. On the other hand, a player who prefers finishing games as quick as possible is less likely to take objects they don't consider necessary.\\
3) ``Common sense'' choices made by humans affect the order in which goals are triggered, especially early in the game, when the player explores the environment. The way players explore the virtual world defines their goals based on the elements they discover. 

IRL algorithms~\cite{Abbeel2004} can identify an expert's goals as a reward function that is learned from demonstrations. However, most of these algorithms assume that the policy demonstrated is optimal for that reward function. From our previous discussion, point 2) can be addressed by such algorithms. For points 1) and 3), an algorithm that can also imitate the behaviour demonstrated is needed for an agent to be able to behave differently when simulating players with different styles, while preserving general human-like decision making. These concepts are known as \textit{intention learning} and \textit{imitation learning}. One method to maximise the advantages of both paradigms is Receding Horizon Inverse Reinforcement Learning (RHIRL)~\cite{MacGlashan2015}.


\section{Background}
\subsection{Interactive Narratives}
\label{sec:ins}
An Interactive Narrative (IN) is a form of storytelling where the listener (reader, player or user) is given freedom to affect the direction of the story by choosing how to act at different stages~\cite{riedl2013interactive}. In some cases, the choices made only affect the order of some events (e.g. the order in which a player asks questions to a Non-Player-Character). However, some of these choices have immediate or delayed consequences (e.g., choosing to join the enemy can cause our current friends to be aggressive instantly, or not asking a specific question can cause a mission to fail after some time because the answer was crucial to finish it).
A modern example of the implementation of an IN is the game \textit{The Witcher 2}~\cite{CDProjectRED}. However, INs  have been around for a long time in different formats, such as the ``Choose your own adventure'' books, or Interactive Fictions (text-based adventure games)~\cite{lebowitz2011interactive}.

\marginpar{%
  \fbox{%
    \begin{minipage}{\marginparwidth}
\textbf{Reinforcement Learning (RL)} consists on finding the best sequence of actions to solve a problem by directly interacting with the environment and observing the outcome of that interaction. RL requires the problem to be modelled as a Markov Decision Process (MPD), which is a tuple $\{S,A,T,R\}$, where $S$ is a set of states that represent the current status of the environment modelled (e.g., current health points, location, etc.), $A$ is a set of actions (e.g., move, talk, attack), $T$ is a set of transition probabilities ${P(s'|s,a)}$ (the probability of reaching state $s \in S$ after taking action $a \in A$, e.g., the probability of having more health points after taking action ``heal''), $R(s,a,s')$ is a reward function that returns the numeric reward of performing action $a$ in state $s$ to reach state $s'$ \cite{Sutton2017}. A reward function is used to encode the goal of the RL agent; desired states typically yield a high reward, while states the agent should avoid yield a low or negative reward.
The goal in RL is to find a policy $\pi : S \rightarrow A$ that maps states to actions and maximises the specified reward function $R$.
  \end{minipage}}}

To provide freedom of choice, INs are typically structured using~\textit{plot points}, i.e., important events in the story that may or may not depend on other events. The author of the IN can establish precedence constraints between plot points without confining the player to a specific path. For instance, opening a locked door has the constraint that the key must be found and collected first, but the player can A) collect the key, talk to a character, start and complete a new mission in a different location, come back and unlock the door, or B) collect the key and immediately unlock the door.

As the options available to the player grow, the number of possible paths between collecting the key and unlocking the door from our example grows as well. If we consider the possible paths between the beginning and the end of the game, picking one that exactly matches the path discovered by a specific player becomes a challenging task. This comes down to making the same decisions as the player in question.
The appeal of INs lies partially on their re-playability. Depending on the plot constraints established by the IN's author, it is possible for the player to reach the end without having discovered all the events and items in the IN's world. In fact, some INs are designed with more than one ending. This may encourage some players to repeat the game to find out what happens if they choose different actions. While this flexibility is an advantage, there are risks of players getting lost in the game by not being able to meet plot point constraints, or \textit{plot gaps} in the resulting story because they discovered plot points relevant to an ending different to the one they reached.

Having a way to simulate the behaviour of different players can help authors identify issues in the IN before launching the product, or help a drama manager (an artificial game master embedded in the game) make decisions in the story during play time to personalise the player's experience~\cite{riedl2013interactive}.
A player model can be used to simulate the decision-making process of players, provided it encodes information representative of the player and makes good use of it. Player modelling is the use of computational intelligence to build models of players interaction with games~\cite{Yannakakis2013}. Player models can include player profiles, a collection of static information not necessarily related to the game, such as cultural background, gender, age, etc.


    

\subsection{Apprenticeship Learning}
\label{sec:irl}
 When a reward function is not clear, or it is difficult to specify as part of an MDP to derive a policy using RL, it is possible to learn a reward function from an expert's demonstrations, to then generate a policy that maximises such reward function. In this case, the ``true'' reward function is represented as a weighed vector of features $R^*(s)=w^*\cdot\phi(s)$, where $||w||_1^*\leq1$ and the vector $\phi: S \rightarrow[0,1]^k$ contains features indicating the trade-off between the desiderata of the task. For example, when learning how to drive, $\phi$ can contain features such as lanes, collisions or off-road.  The process of learning from demonstrations is called Apprenticeship Learning (AL), and formally speaking, the process of learning a reward function from policies available is referred to as Inverse Reinforcement Learning (IRL)\cite{Abbeel2004,ng2000algorithms}.

\subsection{Receding Horizon IRL}
\label{sec:rhirl}
Receding Horizon IRL (RHIRL) is a model-based method that aims to match expert behaviour using a Receding Horizon Controller (RHC) that allows to approximate a value function for a finite horizon of $h$ steps into the future. After each decision, $h$ decreases by 1~\cite{MacGlashan2015}. RHIRL is based on Maximum Likelihood IRL (MLIRL)~\cite{Babes-Vroman2011}. Given a set of demonstrations $D$, the likelihood of $D$ given reward function $R$ is defined as $L(D|R) = \prod_{t \in D} \prod_i^{|t|} \pi_{R}(s_i,a_i)$, where $\pi_R(s_i,a_i)$ is a Boltzman policy to define the probability of taking action $a$ in state $s$ when the reward function to maximise is $R$: $\pi(s,a) = \frac{e^{\beta Q(s,a)}}{\sum_{a' \in A} e^{\beta Q(s,a')}}$, where $\beta$ is used to indicate how noisy the policy is (i.e., how far from optimal the expert's behaviour is). Large values for $\beta$ will make the policy select the action with the highest value, while $\beta=0$ will select actions randomly.


\marginpar{%
  \fbox{%
    \begin{minipage}{\marginparwidth}
\textbf{State Representation}\\
A state consists of the following structure:\\
- Current location\\
- Locations available\\
- Details for each object in the IN world\\
	  \hspace{6pt}- Its name\\
	  \hspace{6pt}- Whether its is locked\\
	  \hspace{6pt}- Whether it is open\\
	  \hspace{6pt}- Whether it is empty\\
	  \hspace{6pt}- Its contents\\
	  \hspace{6pt}- Whether it can be opened\\
	  \hspace{6pt}- Whether it can be taken\\
	  \hspace{6pt}- Whether it is visible\\
	  \hspace{6pt}- Whether it has been seen \\
- Objects in the inventory\\
- For each character:\\
\hspace{6pt}- Their name\\
\hspace{6pt}- Whether the character is visible\\
- For each plot point:\\
\hspace{6pt}- Its name\\
\hspace{6pt}- Whether it has been visited\\
- For each conversation topic:\\
\hspace{6pt}- Its name\\
\hspace{6pt}- Whether it is known by the player\\
\hspace{6pt}- Whether it has been mentioned\\

 \end{minipage}}}

\section{Methodology}
%

Using player traces collected in our previous work~\cite{Rivera-Villicana2018}, our overall methodology consists of the following steps: 
1){Convert player traces traces to demonstrations for RHIRL}, 
2){Extract policies resulting from RHIRL}, 
3){Execute policies in the Anchorhead engine}. 
The Anchorhead engine is based on the work in~\cite{sharma2010drama}, using their code with permission. 
The RHIRL model was implemented using the Brown-UMBC Reinforcement Learning and Planning (BURLAP) Java library~\cite{MacGlashan2015a}. This implementation required to model Anchorhead as an MDP, i.e., using a state representation, action representation and a transition model.

\subsection{Action representation and transition model}
We embedded the mechanics of the Anchorhead engine in the MDP module to perform the learning tasks offline. We modelled nine parameterised actions, described in table~\ref{table:actions}. The applicability of each action is checked before its execution considering the current state and the parameter received. 

\begin{table}[]
\centering
\caption{Actions for Anchorhead's RHIRL model}
\label{table:actions}
\resizebox{\columnwidth}{!}{%
\begin{tabular}{lll}
  \hline
  Action  & Parameters (Type)                 & Conditioins                                   \\ \hline
  goto    & p (INPlace)                       & p is adjacent to current location.              \\
  examine & o (INObject)                      & o is visible.                                   \\
  take    & o (INObject)                      & o is visible.                                   \\
  use     & o (INObject)                      & o is visible.                                   \\
  unlock  & o (INObject), k (INObject)        & o is visible and k is in inventory.             \\
  open    & o (INObject)                      & o is visible and o is unlocked.                 \\
  say     & t (INTopic)                       & t is known and one character is visible.        \\
  buy     & o (INObject)                      & One character is visible.                       \\
  give    & o (INObject)                      & o is in inventory and one character is visible. \\ \hline
\end{tabular}}
\end{table}






\section{Evaluation}
\subsection{Experimental setup}
We trained RHIRL on different groups of demonstrations from a total of 36 traces generated by 23 anonymous players, with the goal to learn one policy per group. We trained on groups rather than all the traces available to learn one single policy because the game traces collected were too noisy for RHIRL to learn a policy that reached an end of the game in this way.
We created groups of game traces to train RHIRL as follows:
\begin{itemize}[leftmargin=*, itemsep=0mm]
\item{By end reached. One group corresponded to End1 (FindEvilGod), and one corresponded to End2 (DiscoverBookInSewers).}
\item{By player profile as defined in our previous work~\cite{Rivera-Villicana2018}. We created two groups for each player profile factor, for low and high scores respectively. In each of these groups, we selected traces that contained similar values for the three PP factors that were not being assessed with the aim of reducing the chances of introducing noise in the behaviour learned. For example: To make the groups of persistent and not persistent players, we only considered the traces where the value scores for familiarity, gaming experience and preference to explore had the same values, and split them into two groups depending on whether the binarised value for persistence was equal to 1 or 0.}
\end{itemize}

In each group, we tested RHIRL with horizons of 1, 2, 3 and 4, each with $\beta$ values of 0.1, 0.5 and 1. By adjusting the values of $h$ and $\beta$, we aim to make RHIRL find a reward function that explains the behaviour observed in demonstrations and generate policies that mimic such behaviour. A horizon $h=0$ is expected to copy the behaviour of the expert, but not generalise well to states not observed in demonstrations. A horizon close to infinite should find the true reward function, allowing better performance in novel states, at the expense of a high computational cost and behaviour less similar to that of the expert.
$\beta$ is useful to specify the optimality of the behaviour observed; a value of $\beta$ close to zero means that the behaviour observed is noisy and sub-optimal, which means that the algorithm will not ``trust'' the expert, therefore, it will try to explore actions rather than imitating them.

\marginpar{%
  \vspace{-16pc}
  \fbox{%
    \begin{minipage}{\marginparwidth}
      \textbf{Player Profile (PP)} \\
      Consists of four variables defined with the aim to differentiate between player styles in an IN~\cite{Rivera-Villicana2018}:
      \begin{itemize}[leftmargin=*, itemsep=0mm]
      \item \textit{Familiarity} with the game being tested (Anchorhead in this case).
      \item \textit{Gaming experience} with any genre.
      \item \textit{Preference to explore} the game's world (e.g, interact with objects or characters, visit nearby locations, etc).      
      \item \textit{Persistence} (e.g., attempting the same quest multiple times).
      \end{itemize}
      
      This player profile was designed to model player behaviour as a function of four variables rather than labeling their behaviour with a stereotype, with the aim of covering a more diverse behaviour range than a defined number of stereotypes. However, in this early work, we only use binarised values.
      The PP of each player was obtained during game data collection and the values for each variable were normalised between 0 and 1.\\
    \end{minipage}}\label{sec:playerprofile} }

In all cases, the maximum number of iterations for RHIRL was set to 10. With the current model, this number of steps using $h$=4 took up to five hours to complete for one group. In future work, we aim to evaluate configurations with more steps and potentially larger horizons.  The policies generated with each combination of $h$ and $\beta$ were recorded. The maximum number of actions recorded in each policy was set to 100, but policies that reached an end of the game usually contain fewer actions. These policies were run on the Anchorhead engine. Following the same approach as in our previous work, we measured the similarity between traces using the Jaccard index of the sets of plot points discovered in each player log and those discovered in each RHIRL policy~\cite{Rivera-Villicana2018}. These trace comparisons were made only between the policy and all the player traces available in each group. For example, the policy obtained from the group of traces that reach End1 is compared to each of the traces belonging to this group, but not to the traces that reached End2 or any of the groups corresponding to factors of the player profile.

\subsection{Results}
\begin{figure}
\centering
\includegraphics[width=\linewidth]{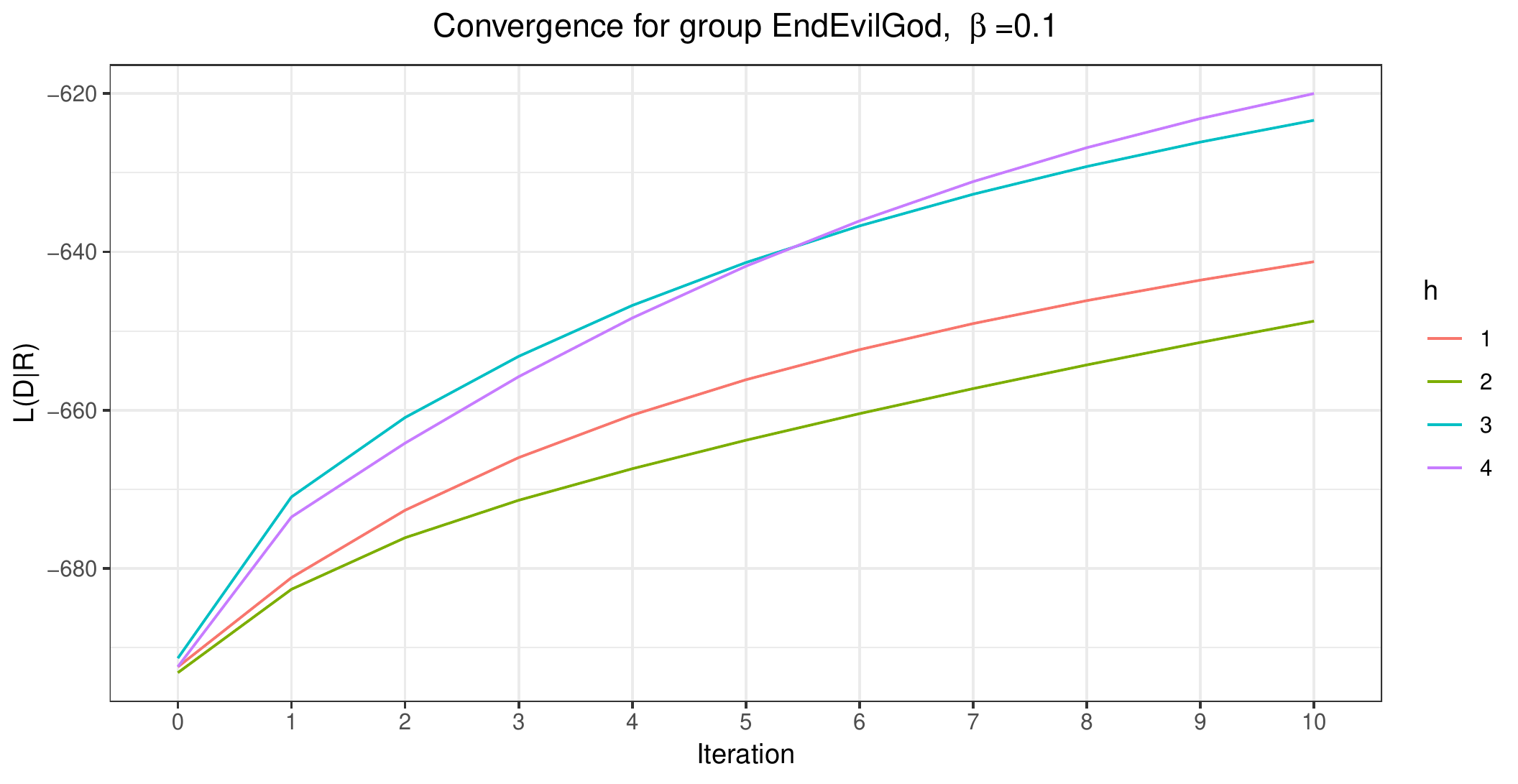} 
	\caption{Convergence of RHIRL for End1}
	\label{fig:convergence_b.1_evilgod}
\end{figure}

Figure~\ref{fig:convergence_b.1_evilgod} shows the convergence of the case where we observed the best performance (i.e., produces behaviour most similar to the group of traces it learned from). This was the group with policies that reached End1 with $\beta=0.1$. As expected from the discussion in the experimental setup, the best performance was observed with the value of $\beta$ that was closest to zero, reflecting the sub-optimality of players' behaviour. The figure shows the progression of the likelihood calculated throughout the 10 steps for the policy learned given the reward function with respect to the different values used for $h$, showing that a larger horizon helps the algorithm converge to $L(D|R)$ faster than shorter horizons. We observed that performance decreased with a horizon of five. This drop in performance is possibly due to additional noise introduced by using demonstrations with a variety of styles.

\begin{table}
  \caption{Similarity results using RHIRL with $h$=4 and $\beta$=0.1} 
  \label{table:irl_summary}
\resizebox{\columnwidth}{!}{%
\begin{tabular}{@{\extracolsep{3pt}} lccccc} 
\hline \\[-1.8ex] 
Group & Mean & Std. Dev & Median & Min & Max \\ 
\hline \\[-1.8ex] 
EndEvilGod & $0.597$ & $0.154$ & $0.545$ & $0.444$ & $0.800$ \\ 
EndSewers & $0.234$ & $0.084$ & $0.269$ & $0$ & $0.304$ \\ 
Experienced & $0.405$ & $0.074$ & $0.400$ & $0.320$ & $0.500$ \\ 
Explorers & $0.303$ & $0.066$ & $0.333$ & $0.227$ & $0.348$ \\ 
Familiar & $0.344$ & $0.037$ & $0.347$ & $0.280$ & $0.423$ \\ 
NonExperienced & $0.303$ & $0.066$ & $0.333$ & $0.227$ & $0.348$ \\ 
NonExplorers & $0.290$ & $0.036$ & $0.300$ & $0.250$ & $0.320$ \\ 
NonFamiliar & $0.394$ & $0.096$ & $0.438$ & $0.269$ & $0.500$ \\ 
NonPersistent & $0.405$ & $0.074$ & $0.400$ & $0.320$ & $0.500$ \\ 
Persistent & $0.231$ & $0.109$ & $0.231$ & $0.154$ & $0.308$ \\ 
\hline \\[-1.8ex] 
\end{tabular}}
\end{table}

Table~\ref{table:irl_summary} shows the results obtained after measuring the similarity of the traces generated by RHIRL versus traces in the corresponding group. Again, the algorithm trained with the group of traces that reached End1 reached the best overall performance.
Our first hypothesis to explain this result is the fact that End1 is easier to reach than End2. Reaching End2 requires the execution of more specific action sequences than End1, which may be mistakenly identified by RHIRL as repeated states despite our state representation considering the plot points reached. Furthermore, the only case where the algorithm produced policies with zero similarity to the real player traces is when using the group of traces that reached End2.

\marginpar{%
  \vspace{-26.5pc}
  \fbox{%
    \begin{minipage}{\marginparwidth}
      \textbf{Related Work} \\
      AL applications related to our work include a Capture The Flag commander that learns policies from encounters with enemies~\cite{Ivanovo2015}, a dungeon generation tool~\cite{Sheffield2018}, and a Super Mario controller that learns action sequences from player demonstrations~\cite{lee2014learning}. While these works apply AL via IRL techniques in a game environment, they do not focus on INs. Their models, unlike ours, assume that optimal demonstrations are provided to the AL algorithm. Related work specifically in INs includes learning player behaviour for believable characters~\cite{Zhao2012}, goal recognition using probabilistic methods~\cite{Baikadi2013, Albrecht1998}, and MDP representations of INs for drama management~\cite{thue2012procedural, Nelson2006a, roberts2006targeting}. These works, although focused on the IN domain, do not implement AL. As for modelling player behaviour, related work includes persona modelling for procedural content generation, where archetypical models of players are used to evaulate playability of generated game content~\cite{Holmgard2014,Liapis2015}. These works focus on game genres other than INs, and their model uses stricter archetypes than ours.
  \end{minipage}}\label{sec:relatedwork}}

We observed a lower convergence rate in groups with a larger amount of traces. Nevertheless, the results indicate that the policies reach higher similarities in such groups. This suggests that the algorithm is able to devise more group-specific policies when learning from a more varied set of demonstrations, which can explain some outliers with similarities of up to 0.666 for the ``Familiar'' group of traces using $h$=3 and $\beta$=0.1. This, however, introduces the problem of finding the optimal number of demonstrations to obtain a working policy in a reasonable amount of steps, which translates in more time required for training.

\section{Conclusion and Future Work}
We have presented an approach to emulate players behaviour in INs using Receding Horizon Inverse Reinforcement Learning (RHIRL), aiming to investigate whether Apprenticeship Learning (AL) could capture human-like decision making in an IN environment, and learn to play an IN in a way that resembles the style of different players. We trained and tested RHIRL with different parameters to control the algorithm's imitation and intention learning rates.

The results suggest that for RHIRL to learn a policy that reaches a fair number of plot points, configurations aiming for intention learning (i.e., a large step horizon and assuming player behaviour is sub-optimal) work better than those aiming for imitation learning (short horizon and assuming player behaviour is optimal). These findings support our hypotheses regarding the sub-optimality of human behaviour in INs, as the best performance was observed using configurations that account for noise in the demonstrations provided. 

In future work, we aim to run a more rigorous training scheme, especially with larger horizons and more player demonstrations. Our first goal is to have an agent learn policies that reach both game endings, followed by an evaluation of the generalisability of our approach on unseen player traces. Findings such as the drop in performance of the RHIRL model with a horizon of 5 are worth further investigation.
Another goal is to develop a hybrid model using RHIRL and more declarative models like the Belief-Desire-Intention approach in our previous work to mitigate the learning difficulties due to the noise in player traces~\cite{Rivera-Villicana2018, Rivera-Villicana2016}.

\bibliographystyle{SIGCHI-Reference-Format}
\bibliography{library}
\end{document}